# Dialectic reasoning with inconsistent information

**Morten Elvang-Gøransson**
Centre for Cognitive Informatics
University of Roskilde
DK-4000 Roskilde, Denmark

**Paul Krause and John Fox**
Advanced Computation Laboratory
Imperial Cancer Research Fund
London WC2A 3PX, UK

## Abstract

From an inconsistent database non-trivial arguments may be constructed both for a proposition, and for the contrary of that proposition. Therefore, inconsistency in a logical database causes uncertainty about which conclusions to accept. This kind of uncertainty is called logical uncertainty. We define a concept of "acceptability", which induces a means for differentiating arguments. The more acceptable an argument, the more confident we are in it. A specific interest is to use the acceptability classes to assign linguistic qualifiers to propositions, such that the qualifier assigned to a propositions reflects its logical uncertainty. A more general interest is to understand how classes of acceptability can be defined for arguments constructed from an inconsistent database, and how this notion of acceptability can be devised to reflect different criteria. Whilst concentrating on the aspects of assigning linguistic qualifiers to propositions, we also indicate the more general significance of the notion of acceptability.

## 1 INTRODUCTION

For classical logic, the presence of an inconsistency in a logical theory is pathological; everything follows from a deduction of falsum, $\bot$. This property of classical logic – and also of intuitionistic and many modal logics – is not, however, a feature which is reflected in "pragmatic" – in the sense of everyday – reasoning. Gabbay and Hunter (1991) argue from a number of cases that people generally have an ability to localize inconsistency, and often suspend the resolution of a contradiction if it does not involve information which is directly relevant to the action at hand. There has been a steady interest in developing models for reasoning in the presence of inconsistent data in both the AI (Dubois, Lang & Prade, 1992; Fox, Krause & Ambler, 1992; Perlis, 1989; Wagner, 1991; Benferhat, Dubois & Prade, 1993) and philosophical logic (Nelson, 1949; Priest, 1989; Priest, Routley & Normann, 1988) communities. Here we will describe a form of dialectic reasoning, in which the presence of arguments both for and against a proposition does not lead to trivialization, but merely affects the "acceptability" of the proposition (and the propositions to which it is related). Our motivation is to understand how certain arguments constructed using classical logic from an inconsistent database can be taken to be more acceptable than others. We want to be able to make such a differentiation purely on the basis of the arguments that can be constructed from a database. The solution we suggest assigns different degrees of acceptability to arguments on the basis of other constructible arguments. We view these different degrees of acceptability as reflecting a kind of uncertainty, which we call logical uncertainty.

To aid the understanding of acceptability as logical uncertainty, a linguistic qualifier is assigned to each of the respective acceptability classes. The particular use of linguistic qualifiers to express uncertainty has been addressed by a number of authors. Most give such terms a semantics in terms of interval valued probabilities (Dubois et al, 1992) or fuzzy sets (Zadeh, 1975). However, Fox (1986) held that such terms were more naturally defined on a qualitative, or symbolic, basis. The advantages of the use of predicates defined on the basis of patterns of argument were demonstrated in a prototype medical decision making application (Fox et al, 1990). In this paper we will offer a set of linguistic qualifiers which are defined on purely logical grounds.

As we worked with the linguistic qualifiers we discovered that the classification we gave of arguments according to their degree of acceptability had a significance beyond the application to the assignment of the linguistic terms. It is possible to reformulate the formalisms defined by various authors using the notion of acceptability. As a specific example, we will consider Poole's notion of specificity (Poole 1985). After having discussed how various degrees of acceptability can be introduced purely on the basis of the constructible arguments, we also consider how the notion of acceptability can be extended to allow additional criteria to be taken into account. As a specific example, we will indicate how explicit priorities can be taken into account. There are several ways in which this can be done, cf. for instance Hunter (1992), and we will just consider one of these.

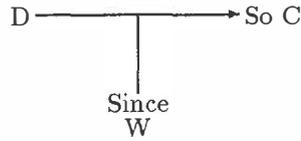

Figure 1: Toulmin's Schema

The consequence relations we introduce for constructing arguments are defined as a Labelled Deductive Systems (LDS), cf. Gabbay (1992). The main idea of LDS is that of labelling formulas and using information stored in these labels. This idea fits well with what we are doing. The idea of using arguments as the fundamental logical entity is inspired by, but not directly based on, work on the development of a "Logic of Argumentation" (Fox et al, 1992; Krause et al, 1993). In this paper we relate our work instead to the philosophical account of arguments offered by Toulmin (1956).

The structure of the main body of the paper is as described above. We start by defining our general model for dialectic reasoning with inconsistent information. This model gives a general definition of what we call "systems of argumentation". Alongside this definition we draw some parallels to existing formalism.

## 2  ARGUMENTATION

We model dialectic reasoning with inconsistent information as argumentation, which we define as the construction and use of arguments. Argumentation is a general principle, which can be instantiated with specific ways of constructing and using arguments. A specific instance of the argumentation principle is called a "system of argumentation". We introduce the general principle of argumentation and explain it through a simple example.

Toulmin (1956) provides an informal model of argumentation, which in it's basic form can be illustrated as in Figure 1 (Toulmin, 1956, p. 99). Informally, Toulmin's "schema" reads as this: "Warranted by the general principles, W, conclusion $C$ can be concluded from the facts, D". The essence of Toulmin's account is that arguments carry information about the facts and warrants from which the conclusion of the argument has been established. For similar reasons we assign labels to arguments, and in our account arguments are modelled as pairs, the first component is the conclusion of the argument and the second component is the label of the argument. The label carries, using Toulmin's terminology, information about the facts and warrants of the argument. For each specific definition of an argument, in a system of argumentation, the label must carry sufficient information for assessing the acceptability of the argument, cf. below.

We model facts as items in a labelled database, where each fact is assigned a unique label. We call such databases "flat", if there is no structure imposed on the labels and "prioritized" otherwise. In the last section we will discuss the use of priorities, but until then we only consider flat databases. The following example indicates how information in a database can be labelled.

**Example of a flat database, called $K_M$:** (Literates will recognize r4 as "modus Montanus" (Holberg).)

$$\begin{aligned}
r1: &\quad mother(x) \rightarrow \neg flies(x) \\
r2: &\quad mother(x) \rightarrow \neg stone(x) \\
r3: &\quad stone(x) \rightarrow \neg flies(x) \\
r4: &\quad q \rightarrow ((p \rightarrow q) \rightarrow p) \\
f1: &\quad mother(Karen)
\end{aligned}$$

Warrants are throughout modelled as rules of classical logic, and they are assigned a passive role in the present account of argumentation.

Having decided what form facts and warrants have, we can define how arguments can be constructed. The "constructible" arguments from a specific database are defined by an "argumentation consequence relation". In the definition of an argumentation consequence relation, it must be made explicit how information about the argument is aggregated in its label.

By way of illustration, we continue the example. Consider the following consequence relation, consisting of two rules. Ax allows for facts in the database to be used and Modus Ponens, $\rightarrow$-E, allows for these facts to be combined. (Ax and $\rightarrow$-E are part of the inference system defined in Figure 2.)

**Example of an arg. cons. rel.:**

$$\text{Ax} \frac{}{K \vdash (p, a)} \ (p, a) \in K$$

$$\rightarrow\text{-E} \frac{K \vdash (p \rightarrow q, a) \quad K \vdash (p, b)}{K \vdash (q, a \cup b)}$$

The definition makes explicit how the labels of the facts in some database, $K$, are propagated in the construction of arguments. Here, arguments are pairs of a formula and a set of labels of facts in the database on which the argument is based. For simplicity we consider facts to be labelled with singleton sets. For instance $(mother(Karen), \{f1\}) \in K_M$.

From the database $K_M$ and the argumentation consequence relation defined above, we can construct the following arguments:

$$(\neg stone(Karen), \{f1, r2\})$$

$$(stone(Karen), \{f1, r1, r3, r4\})$$

$$\text{Ax} \frac{}{K \vdash (p,a)} \ (p,a) \in K \qquad \vee\text{-I1} \frac{K \vdash (q,a)}{K \vdash (p \vee q, a)}$$

$$\top\text{-I} \frac{}{K \vdash (\top, \emptyset)} \qquad \vee\text{-I2} \frac{K \vdash (q,a)}{K \vdash (p \vee q, a)}$$

$$\rightarrow\text{-I} \frac{K, (p, \emptyset) \vdash (q, a)}{K \vdash (p \rightarrow q, a)} \qquad \vee\text{-E} \frac{K \vdash (p \vee q, a) \quad K, (p,a) \vdash (r, b') \quad K, (q,a) \vdash (r, b'')}{K \vdash (r, b' \cup b'')}$$

$$\rightarrow\text{-E} \frac{K \vdash (p \rightarrow q, a) \quad K \vdash (p, b)}{K \vdash (q, a \cup b)} \qquad \neg\text{-I} \frac{K, (\neg p, \emptyset) \vdash (\bot, a)}{K \vdash (p, a)}$$

$$\wedge\text{-I} \frac{K \vdash (p,a) \quad K \vdash (q,b)}{K \vdash (p \wedge q, a \cup b)} \qquad \neg\text{-E} \frac{K \vdash (p,a) \quad K \vdash (\neg p, a)}{K \vdash (\bot, a)}$$

$$\wedge\text{-E1} \frac{K \vdash (p \wedge q, a)}{K \vdash (p, a)} \qquad \text{RAA} \frac{K, (p, \emptyset) \vdash (\bot, a)}{K \vdash (\neg p, a)}$$

$$\wedge\text{-E2} \frac{K \vdash (p \wedge q, a)}{K \vdash (q, a)} \qquad \text{EFQ} \frac{K \vdash (\bot, a)}{K \vdash (p, a)}$$

Figure 2: Argumentation Consequence Relation

Suppose we want to draw a conclusion from this set, called $A_M$, of arguments. Before we can do so we must agree on a policy for drawing such conclusions, and we then define such a policy as a flattening function (the terminology is due to Gabbay 1992). In the case of the above example, we have decided to allow arguments to be based on any fact in the database apart from "modus Montanus".

**Example of a simple flattening function:** Let $A$ be any set of arguments. Then:

$$\text{Flat}(A) = \{p \mid (\exists a)((p,a) \in A \wedge r4 \notin a)\}$$

Therefore, the result of flattening the above two arguments,

$$\text{Flat}(A_M) = \{\neg stone(Karen)\},$$

reveals that Mother Karen is not made of stone. This policy is indeed very simple and specific for the example we have given.

So far a system of argumentation is nothing but a LDS and everything we have done is in the realm of the general definitions that Gabbay (1992) gives. We will now specialize our framework towards handling inconsistency by formalizing the notion of acceptability. This notion appears to be fundamental for the uses of argument to handle logical uncertainty. It will be used here for making uniform definitions of flattening functions that reveal the logical uncertainty inherent in a set of arguments.

Before proceeding with this, we will recall Toulmin's account of this problem. According to Toulmin, an argument can be represented as a conclusion together with with information about the facts and warrants from which the argument can be constructed. Presented with such an argument, doubts may be raised in either its conclusion or in the facts and warrants supporting the conclusion. If sufficiently convincing arguments can be constructed for doubt in the conclusion of an argument, the argument is said to be "rebutted". If, on the other hand, convincing arguments can be constructed for doubt in the facts or warrants from which an argument has been constructed, then the argument is said to have been "undercut". This defines, in principle, two notions of defeat which are common in the AI literature. cf. for example, Loui's notion of defeasible arguments (Loui, 1987), Nute's (1988) and Pollock's (1992) models of defeasible reasoning. However, in all these three cases propositions can only be assigned to one of the classes true or false. We wish to assign a finer grading than just truth and falsity, which better reflects the logical certainty of a proposition.

The approach we take is to define classes of acceptability for constructible arguments. Such classes are called "acceptability classes" and they can be defined for any argumentation consequence relation. Some of the defined classes will be counted as more acceptable than others. This induces an "acceptability ordering" over arguments, defining different discrete "degrees of acceptability" that an argument can have. The "acceptability of an argument" is defined as its maximal degree(s) of acceptability if any such can be defined. Arguments of the same degree of acceptability are intented to have the same logical certainty. A specific acceptability class is defined relative to other classes of arguments as well as by the use of some absolute requirements. An acceptability class can be conceived as the set of all those arguments from some set (the set of defining arguments) that are able to pay the price for membership. This price consists of two parts, each of which must be settled:

- an absolute requirement, and
- a requirement relative to some set of arguments (the set of moderating arguments).

The notion of acceptability induces a flattening policy, by picking the most acceptable arguments. This provides a firm basis for imposing (non-logical) heuristics for resolving inconsistencies and making decisions, and it allows for the introduction of uncertainty measures to assert varying degrees of acceptability. Our main example, which occupies the rest of this paper, will further clarify these remarks and also the vague terms in which the whole notion of acceptability has been introduced.

Preliminary investigations have shown that instances of the proposed framework embrace several formal systems. We already mentioned three above, and all of these appear to be reexpressible in terms of acceptability. As one specific example, we will argue that Poole's notion of specificity is a specific instance of acceptability. A similar argument can be constructed for the work of Wagner (1991). [1]

**Specificity as acceptability:** Poole (1985) has an argument-like notion called "explanation". An explanation is constructed using classical entailment from contingent facts together with a set of necessary facts and hypotheses. Specificity corresponds to the minimal set of contingent facts (required for some set of hypotheses to participate in giving an explanation) being of a certain "size", i.e. the larger the more specific, and this induces a specificity ordering among arguments. The notion of being most specific is relative to other arguments. Consider as an example the set of hypotheses: $\{p \to q, p \land r \to \neg q\}$, the set of necessary facts $\emptyset$ and the set of contingent facts $\{p, r\}$. From this set of hypotheses and facts, using Poole's definition we can construct a minimal argument $\{p, r, p \land r \to \neg q\} \vdash \neg q$, for $\neg q$ which is more specific than the minimal argument $\{p, p \to q\} \vdash q$, which we can construct for $q$. Hence $\neg q$ is the more acceptable claim in this context. Specificity is a notion of acceptability defined using logical as well as non-logical means. The non-logical part stems from the delimitation of the necessary facts.

We summarize this section by making precise what a system of argumentation is.

**System of Argumentation:** A system of argumentation is an argumentation consequence relation and a flattening function induced by a notion of acceptability. The argumentation consequence relation describes how new arguments can be constructed from a database and a set of warrants. Arguments carry labels with information about their support. The flattening function defines how conclusions are selected from a set of arguments, using a notion of acceptability that has been designed to reflect the information that is available about individual arguments.

This definition is a quite general specialization of the notion of a LDS, which as argued above, fits in with many existing formalisms.

In the remaining sections we concentrate on defining a system of argumentation that assigns linguistic qualifiers to arguments constructed from an inconsistent database.

## 3  CONSTRUCTING ARGUMENTS

We define an argumentation consequence relation, where formulas are labelled with the names of the facts from which they have been derived (just as in the previous example).

**Database:** A database, $K$, is any, consistent or inconsistent, set of uniquely named propositions. If $(p, \{l\}) \in K$, where $l$ is labelling the proposition $p$, then $K(l) = p$.

For simplicity we assume that there is a one-one correspondence between fact names and facts in any database, and it therefore makes sense to refer to the (set of) fact(s) labelled by a (set of) label(s).

**Argument:** Let $p$ be a proposition and $a$ a set of fact names. Then $(p, a)$ is an argument for $p$ supported by $a$, iff $a$ is a minimal set of labels, such that:

$$K \vdash (p, a).$$

The argumentation consequence relation is defined in Figure 2.

**Non-trivial argument:** An argument $(p, a)$ is non-trivial if the set of facts labelled by $a$ is consistent.

**Tautological argument:** An argument $(p, a)$ is tautological if $a = \emptyset$.

**Defeat:** Let $(p, a)$ and $(q, b)$ be arguments from $K$. The argument $(q, b)$ can be defeated in one of two ways. Firstly, $(p, a)$ "rebuts" $(q, b)$ if $p \to \neg q$. Secondly, $(p, a)$ "undercuts" $(q, b)$ if for some $l \in b$, labelling a fact $r$, $p \to \neg r$.

## 4  ACCEPTABILITY CLASSES

We may now define a hierarchy of acceptability classes using the logical notions of defeat and argument. The classes defined in Figure 3 reflect increasing degrees of acceptability, for arguments constructible from any database $K$. We can now further clarify our distinction between relative and absolute membership criteria. The absolute criteria for membership of $A_1$, $A_2$

---

[1] Lately, we realised that our views appear, especially as formulated in an earlier paper Elvang, Krause & Fox (1993), to coincide closely with those of Pinkas & Loui (1992) and that their "cautiousness" is similar to our acceptability.

$$A_1(K) = \{(p,a)|(p,a) \text{ is an argument from } K\}$$

$$A_2(K) = \{(p,a) \in A_1(K)|(p,a) \text{ is non-trivial}\}$$

$$A_3(K) = \{(p,a) \in A_2(K)|\neg(\exists b)((\neg p, b) \in A_2(K))\}$$

$$A_4(K) = \{(p,a) \in A_3(K)|(\forall l \in a)((\neg(\exists b)((\neg K(l), b) \in A_2(K)))\}$$

$$A_5(K) = \{(p,a) \in A_4(K)|(p,a) \text{ is a tautological}\}$$

Figure 3: Acceptability Classes

and $A_5$ are respectively that the arguments; simply exist, are consistent, and are tautological. For each class the relative criteria include membership of the previous class (if any). In addition, for $A_3$ and $A_4$ the relative criteria also include the notions of rebutting and undercutting defeat, respectively. The acceptability classes have the following relationships

**Properties:**
$A_5(K) \subseteq A_4(K) \subseteq A_3(K) \subseteq A_2(K) \subseteq A_1(K)$

The relation "more acceptable than" between arguments is defined using the ordering that is induced by the set inclusion hierarchy of the acceptability classes:

**More acceptable than:** Let $(p,a)$ and $(q,b)$ be arguments from $K$. Then the argument $(p,a)$ is more acceptable (w.r.t. $K$) than the argument $(q,b)$, iff for some $i$, $1 \leq i \leq 5$, $(p,a) \in A_i(K)$ and $(q,b) \notin A_i(K)$. If $p,q$ are conclusions in arguments from $K$, then we say that $p$ is more acceptable than $q$ if $p$ has an argument that is more acceptable than any argument for $q$.

This hierarchy can be used as a basis for assigning qualifiers to propositions in such a way that their "logical certainty" is reflected by these terms.

## 5  LINGUISTIC QUALIFIERS

We now assign linguistic qualifiers to arguments of varying degrees of acceptability. Any database, $K$, can be partitioned as defined in Figure 4, and this partitioning defines an assignment of linguistic qualifiers to the arguments that can be constructed from $K$. We understand the words "supported", ... and "certain" in Figure 4 to denote increasing certainty. For instance, probable($K$) contains all constructible arguments from $K$, that are at least plausible.

The subset-ordering over the acceptability classes defined in Figure 3, induces an acceptability relation over arguments, where "certain" is regarded as the best linguistic qualifier. Based on the assignment of linguistic qualifiers to arguments, we can define a flattening function, assigning the best linguistic qualifier to propositions that are the conclusions of some argument. The flattening function is defined over the

$$\text{supported}(K) = A_1(K)$$
$$\text{plausible}(K) = A_2(K)$$
$$\text{probable}(K) = A_3(K)$$
$$\text{confirmed}(K) = A_4(K)$$
$$\text{certain}(K) = A_5(K)$$

Figure 4: assignment of linguistic qualifiers

set of all constructible arguments from some database $K$, and assigns qualifiers to propositions according to the criteria for assigning "basic" linguistic qualifiers to propositions, defined in Figure 5.

Using the basic qualifiers defined in Figure 5, we can also define "hybrid" qualifiers as exemplified in Figure 6. Many more than these can in principle be defined, and a fuller vocabulary is considered in Fox (1986). However, in this paper we do not want to push the natural language analogies too far, and for some of the above suggestions we have clearly not quite captured the "common sense understanding" of the terms. For instance, $implausible(p)$ might be better defined as $plausible(\neg p)$.

**Example:** Let $K$ be the database, labelled as follows:

$f1: \quad p \qquad\qquad r1: \quad p \rightarrow q$
$f2: \quad \neg q \qquad\qquad r2: \quad q \rightarrow r$
$f3: \quad s$
$f4: \quad \neg p$

(This database is similar to $KB_1$ in (Wagner 1991).) We will consider the acceptability of the arguments:

1. $(p, \{f1\})$.
2. $(s, \{f3\})$.
3. $(r, \{r1, r2, f1\})$.
4. $(\bot, \{f1, f4\})$.
5. $(\neg s, \{f1, f4\})$.
6. $(p \rightarrow r, \{r1, r2\})$.

$$supported(p) \quad \text{iff} \quad (\exists a)((p,a) \in \text{supported}(K) - \text{plausible}(K))$$

$$plausible(p) \quad \text{iff} \quad (\exists a)((p,a) \in \text{plausible}(K) - \text{probable}(K))$$

$$probable(p) \quad \text{iff} \quad (\exists a)((p,a) \in \text{probable}(K) - \text{confirmed}(K))$$

$$confirmed(p) \quad \text{iff} \quad (\exists a)((p,a) \in \text{confirmed}(K) - \text{certain}(K))$$

$$certain(p) \quad \text{iff} \quad (\exists a)((p,a) \in \text{certain}(K))$$

Figure 5: Basic Linguistic qualifiers

Argument (1) is plausible, because its conclusion is a fact. (1) is not probable, because a rebutting argument can be constructed using the fact $f4$. (Since $p$ can at the best be given a plausible argument, we have $plausible(p)$.) Argument (2) is confirmed, because no rebutting or undercutting arguments can be constructed. It is interesting to note that the inconsistency of $K$ does not affect the acceptability of (2). Argument (3) is plausible, because no rebutter can be constructed, but (3) is undercut by an argument for $\neg p$. Arguments (4) and (5) are supported, but by definition not plausible. Argument (6) is probable, but not confirmed, because $(\neg(p \rightarrow q), \{f1, f2\})$ is a plausible argument that undercuts (6).

The above example reveals some interesting properties, which explicate how inconsistency in a database can be transformed into uncertainty about the answers that the database can give to queries.

**Properties:** Suppose we have a database that can be disjointly partitioned as:

$$K \cup K1 \cup K2$$

and that $K \cup K1$ and $K \cup K2$ are consistent, but $K1 \cup K2$ is inconsistent. Then we have: Any argument constructible from

- $K$ will be confirmed,
- $K \cup K1$ (or $K \cup K2$) will at least be plausible, and
- $K1 \cup K2$ will be at least supported.

## 6 USING PRIORITIES

So far we have only been concerned with flat databases, where each piece of information is considered to be equally good. In this section we will consider how explicit priorities between facts can be used to define the acceptability of arguments. Priorities need not be given for all the information of a database, but can be limited to what we will call the "focus set", $F$. The set of labels of a database is then partitioned into a focus set and a "background set", $B$. The priorities, defined as a partial order $>$, over the focus set induces a partial ordering over the whole database as follows.

| | | |
|---|---|---|
| $opposed(p)$ | iff | $supported(\neg p)$ |
| $doubted(p)$ | iff | $plausible(\neg p)$ |
| $dubious(p)$ | iff | $probable(\neg p)$ |
| $rejected(p)$ | iff | $confirmed(\neg p)$ |
| $impossible(p)$ | iff | $certain(\neg p)$ |
| | | |
| $implausible(p)$ | iff | $\neg plausible(p)$ |
| $improbable(p)$ | iff | $\neg probable(p)$ |
| $unconfirmed$ | iff | $\neg confirmed(p)$ |
| $uncertain(p)$ | iff | $\neg certain(p)$ |

$equivocal(p)$ iff $supported(p) \land supported(\neg p)$
$problematic(p)$ iff $plausible(p) \land plausible(\neg p)$

Figure 6: Hybrid Linguistic Qualifiers

For some database with labels $F \cup B$, the partial order, $\succ$, is induced:

| if $l, m \in F, l > m$ | then | $l \succ m$ |
| if $l \in F, m \in B$ | then | $l \succ m$ |
| if $l, m \in B$ | then | $l = m$ |

no other items are related

For some database with an induced partial ordering, $\succ$ and arguments, $(p, a)$ and $(q, b)$, the priority of $(p, a)$ over $(q, b)$, $(p, a) \succ_p (q, b)$, is defined as:

$$(\exists l \in a, \forall m \in b)(l \succ m).$$

Respect can be paid to the priorities, by changing the definition of probable, cf. the acceptability class, $A_3$. The refined definition of this class is:

$A_3(K) =$
$\quad \{(p, a) \in A_2(K) |$
$\quad \neg(\exists b)((\neg p, b) \in A_2(K) \land (p, a) \not\succ_p (\neg p, b))\}$

We will show by use of an example how the priorities over the focus set can be extended to a partial order over the full set of labels of a database, and how this affects the conclusions that can be drawn.

Suppose we have the following database:

| $f1:$ | $i$ | $r1:$ | $gu \rightarrow \neg du$ |
| $f2:$ | $d$ | $r2:$ | $i \rightarrow gu$ |
| | | $r3:$ | $d \rightarrow du$ |

The database represents a doctor's conception of the status of his patient, who complains of pain in the stomach. The patient explains that on different occasions he has both what he considers as immediate ($i$) and delayed ($d$) stomach pain after meals, but that the immediate pain is more dominant than the delayed. This defines the doctor's focus set as:

$$\{f1, f2\}$$

with the additional information that:

$$f1 > f2.$$

From past experience, the doctor knows that immediate stomach pain after meals is an indicator of gastric ulcer ($gu$) and that delayed pain is an indicator for duodenal ulcer ($du$). The doctor's experience also counts it as unlikely for these two diseases to occur simultaneously. Therefore, the doctor's background set is:

$$\{r1, r2, r3\}$$

Using the acceptability classes defined in the previous section, i.e. without taking the dominance of ($i$) into account, we have a situation where any of the propositions $du, \neg du, gu, \neg gu$ can at the very best be given a plausible argument. Neither of them have a probable argument. Therefore there is a conflict: neither is more acceptable than another.

For the database above the partial order:

$$f1 \succ f2 \succ r1 = r2 = r3,$$

is induced. According to this definition, the argument $(\neg du, \{f1, r1, r2\})$ for $\neg du$ has higher acceptability than the argument $(du, \{f2, r3\})$ for $du$. Similarly $(gu, \{f1, r2\})$ is of higher acceptability than $(\neg gu, \{f2, r1, r3\})$.

Using the changed definitions, the doctor will be able to confirm for herself, that gastric ulcer is the most acceptable explanation of the patients symptoms.

## 7  Final remarks

Two different conclusions can be drawn from this paper. First regarding the notion of acceptability, which we suggested as a tool aiding the resolution of conflicts arising from logical uncertainty. We think that the idea of classifying arguments according to their acceptability offers an interesting formalization of dialectic reasoning. We find it particular interesting that notions of acceptability appear to be implicit in many existing formalisms and hope that this new view on logical uncertainty can add further insight.

Our conclusion regarding the assignment of linguistic qualifiers to acceptability classes is more soft. In discussing the work on linguistic qualifiers in this paper with colleagues, we have often described it as an "interesting experiment" in reasoning under uncertainty. That seems to be a fair assessment of its current status. We are not suggesting that this work be taken as a serious suggestion for anything like a natural language semantics, although it is our view that some of the natural language usage of the linguistic terms that we introduce have been covered. If the terms are then used in combination with more sophisticated systems of argumentation, like the one taking explicit priorities into account, then this may well provide sufficient discriminatory power for many applications in decision support. This will be especially useful in those domains where the elicitation of reliable numerical uncertainty coefficients cannot be guaranteed. Models of uncertain reasoning based on a qualitative evaluation of arguments have been shown to perform effectively (Chard, 1991; O'Neil & Glowinski, 1990). Providing a more formal basis for such models will help in defining their properties, and in their further refinement, so this work does raise some exciting possibilities.

## Acknowledgement

Paul Krause is supported under the DTI/SERC project 1822: a Formal Basis for Decision Support Systems. This work was carried out whilst Morten Elvang-Gøransson was a guest worker at The Imperial Cancer Research Fund and he would like to thank the ICRF for having granted access to office facilities during 1992/93. The authors are thankful to the anonymous referees and Dr. Anthony Hunter.